\documentclass[conference,10pt]{IEEEtran}

\ifCLASSINFOpdf
  \usepackage[pdftex,final]{graphicx}
  \usepackage{epstopdf}
  \graphicspath{ {./images/} }
  \DeclareGraphicsExtensions{.pdf,.jpeg,.png}
\else
\fi

\usepackage{bm}
\usepackage{dsfont}
\usepackage{amssymb,amsmath}
\usepackage{mathtools}

\usepackage{subcaption}

\usepackage[hyphens]{url}

\newcommand{\RR}{\mathbb{R}}

\begin{document}

\title{One-Class SVM with Privileged Information and its Application to Malware Detection}

 \author{
 \IEEEauthorblockN{Evgeny Burnaev}
 \IEEEauthorblockA{Skolkovo Institute of Science and Technology,\\
Building 3, Nobel st., Moscow 143026, Russia}
 \IEEEauthorblockA{Institute for Information Transmission Problems,\\
19 Bolshoy Karetny Lane, Moscow 127994, Russia,\\
 Email: e.burnaev@skoltech.ru}\\
 \and
 \IEEEauthorblockN{Dmitry Smolyakov}
 \IEEEauthorblockA{Institute for Information Transmission Problems,\\
 19 Bolshoy Karetny Lane, Moscow 127994, Russia,\\
 Email: dmitry.smolyakov@iitp.ru}
 }

\maketitle

\begin{abstract}
A number of important applied problems in engineering, finance and medicine can be formulated as a problem of anomaly detection based on a one-class classification. A classical approach to this problem is to describe a normal state using a one-class support vector machine. Then to detect anomalies we quantify a distance from a new observation to the constructed description of the normal class. In this paper we present a new approach to one-class classification. We formulate a new problem statement and a corresponding algorithm that allow taking into account privileged information during the training phase. We evaluate performance of the proposed approach using synthetic datasets, as well as the publicly available Microsoft Malware Classification Challenge dataset.
\end{abstract}

\IEEEpeerreviewmaketitle

\section{Introduction}
\label{introduction}

\label{sec:introduction}

Anomaly detection refers to the problem of finding patterns in data that do not
conform to an expected behaviour. Anomaly detection finds extensive use in a wide variety of applications such as fraud detection for credit cards, insurance or health care, intrusion detection for cyber-security, fault detection in safety critical systems, and military surveillance for enemy activities \cite{ocsvmdoc,odappl,odappl2}. A classical approach to anomaly detection is to describe expected (``normal'') behaviour using one-class classification techniques, i.e. to construct a description of a ``normal'' state using a number of examples, e.g. by describing a geometrical place of training patterns in a feature space. If a new test pattern does not belong to the ``normal'' class then we consider it to be anomalous.

To construct a ``normal'' domain we can use well-known approaches such as the Support Vector Domain Description (SVDD) \cite{svdd}, \cite{svddr} and the One-Class Support Vector Machine (One-Class SVM) \cite{ocsvm}, possibly combined with model selection for anomaly detection \cite{MSAD}, resampling \cite{Resampling}, ensembling of ``weak'' anomaly detectors \cite{ensemble} and extraction of important features using manifold learning methods \cite{DR,DR2}. Both SVDD and One-Class SVM can be kernelized to describe a complex nonlinear ``normal'' class.

For the original two-class Support Vector Machine \cite{nusvm} Vapnik recently proposed a modification that allows taking into account privileged information during the training phase to improve a classification accuracy \cite{LUPI}. Let us provide some examples of privileged information. If we solve an image classification problem, then as privileged information we can use a textual image description. In case of a malware detection we can use a source code of a malware to get additional features for the classification. Such information is not available during the test phase (e.g. it could be computationally prohibitive or too costly to obtain), when we use the trained model for anomaly detection and classification, but can be used during the training phase.

In this work we combine these two concepts (SVMs and learning using privileged information) and propose new approaches to train SVDD and One-Class SVM with privileged information, under the intuition that this additional information available at the training time can be utilized to better define the ``normal'' state. Through experiments on synthetic data and some real data from the Malware Classification Challenge (see \cite{malwa}), we show that a model, trained with privileged information, performs better than a model, trained without privileged information. However, we do not want to claim that the used setup of malware classification experiments indeed fully reflects a specificity of cyber security applications. Rather we demonstrate that one-class classification with privileged information is useful for cyber security problems.

\section{One-Class SVM and SVDD}
\label{sec: oneclass}

Below we briefly describe two classical approaches to one-class classification. We are given an i.i.d. sample patterns $(x_1,\ldots, x_l)\in\mathcal{X}\subset\RR^n$. The main idea of these algorithms is to separate a major part of sample patterns, considered to be ``normal'', from those ones, considered to be ``abnormal'' in some sense.
\subsection{One-Class SVM}
\label{sec: oneclasssvm}

In case of the original One-Class SVM \cite{ocsvm} we consider those patterns to be abnormal, which are close to origin of coordinates in a feature space.

Let us separate patterns using a hyperplane in a feature space defined by some feature map $\phi(\cdot)$ and a normal vector to the hyperplane $w$. We consider that a pattern $x$ belongs to a ``normal'' class if $(w\cdot \phi(x)) > \rho$. In order to define the hyperplane, i.e. the normal vector $w$ and the value of $\rho$, we solve an optimization problem
\begin{eqnarray}\label{oneclass}
\frac{1}{2}\|w\|^2_{\ell_2} + \frac{1}{\nu l}\sum\limits_{i=1}^l\xi_i - \rho\to \min\limits_{w, \xi, \rho}  \\
s.t.\,\,\,\,\,\,(w\cdot \phi(x_i)) \geq \rho - \xi_i \label{ocsvm},\,\,\,\,\,\,\xi_i \geq 0.\notag
\end{eqnarray}
Here $\nu$ is a regularization coefficient, $\xi_i$ is a slack variable for the $i$-th pattern.

Optimization problem \eqref{oneclass} is convex, therefore its solution coincides with that of the dual one: \begin{eqnarray}
-\sum\limits_{i=1}^l\sum\limits_{j=1}^l \alpha_i\alpha_j K(x_i, x_j)\to \max\limits_{\alpha} \label{prstr}\\
s.t.\,\,\,\,\,\,\sum\limits_{i=1}^l \alpha_i = 1,\,\,\,\,\,\,0 \leq \alpha_i \leq \frac{1}{\nu l}.\notag
\end{eqnarray}
Here the scalar product
 $(\phi(x_i)\cdot \phi(x_j))$ is replaced by the corresponding kernel function $K(x_i, x_j)$. Therefore as usual we do not need to know an explicit representation of $\phi(x)$ in order to solve the dual problem. Moreover, the solution of the primal problem can be represented through the solution of the dual problem, namely
$w = \sum\limits_{i=1}^l\alpha_i \phi(x_i).$

Thanks to the structure of \eqref{prstr} if some $\alpha_i > 0$, then the pattern $x_i$ belongs to the boundary of the ``normal'' domain, i.e. $(w\cdot \phi(x_i)) = \rho$ \cite{ocsvm}. Thus the offset $\rho$ can be recovered by exploiting the fact that for any $\alpha_i>0$ the corresponding pattern $x_i$ satisfies the equality
\[
\rho = (w\cdot \phi(x_i)) = \sum_{j=1}^{l} \alpha_j K(x_j, x_i).
\]
As a result the corresponding decision rule has the form
\[
f(x) = \sum\limits_{i=1}^l\alpha_i K(x_i, x) - \rho.
\]
In case $f(x)>0$ a pattern $x$ is considered to belong to the ``normal'' class and vice versa. 
\subsection{SVDD}

Another approach to anomaly detection is to separate outlying patterns using a sphere \cite{svdd}, \cite{svddr}. As before we denote by $\phi(\cdot)$ some feature map. Let $a$ be some point in the image of the feature map and $R$ be some positive value. We consider that a pattern $x$ belongs to a ``normal'' class, if it is located inside the sphere $\|a - \phi(x)\|^2_{\ell_2} \leq R$. In order to find the center $a$ and the radius $R$ we solve the optimization problem
\begin{eqnarray}\label{svddopt}
R + \frac{1}{\nu l} \sum\limits_{i=1}^l\xi_i \to \min\limits_{R, a, \xi_i}\\
s.t.\,\,\,\,\,\,\|\phi(x_i) - a\|^2_{\ell_2} \leq R + \xi_i,\,\,\,\,\,\,\xi_i \geq 0.\notag
\end{eqnarray}
Here $\xi_i$ is a distance from the pattern $x_i$, located out of the sphere, to the surface of the sphere. On the face of it, the variable $R$ can be considered as a radius only if we require its positivity. However, it can be easily proved that this condition is automatically fulfilled if $\nu \in (0, 1)$ \cite{svddr}, and for $\nu \not\in (0, 1)$ the solution of \eqref{svddopt} is degenerate.

The dual problem has the form
\begin{eqnarray}
 \sum\limits_{i=1}^l \alpha_i K(x_i, x_i) - \sum\limits_{i=1}^l\sum\limits_{j=1}^l\alpha_i\alpha_jK(x_i, x_j)\to \max\limits_{\alpha}\notag\\
s.t.\,\,\,\,\,\,0 \leq \alpha_i \leq \frac{1}{\nu l},\,\,\,\,\,\,\sum\limits_{i=1}^l\alpha_i = 1.\notag
\end{eqnarray}
As in the previous case here we replace the scalar product $(\phi(x_i)\cdot \phi(x_j))$ with the corresponding kernel $K(x_i, x_j)$.

We can write out the solution of the primal problem using the solution of the dual problem
\[
a = \sum\limits_{i=1}^l \alpha_i \phi(x_i),\,\,\,\,\,\,
R =\|\phi(x_j)\|^2_{\ell_2} - 2 (a\cdot\phi(x_j)) + \|a\|^2_{\ell_2}, \]
where in order to calculate $R$ we can use any $x_j$, such that $\alpha_j>0$. Here $\|\phi(x)\|^2_{\ell_2} = K(x, x)$, $(\phi(x)\cdot a) = \sum_{i=1}^l\alpha_i K(x_i, x)$ and $\|a\|^2_{\ell_2} = \sum_{i=1}^l\sum_{j=1}^l\alpha_i\alpha_jK(x_i,x_j)$.

The decision function has the form
\[
f(x) = K(x, x) - 2\sum\limits_{i=1}^l\alpha_i K(x, x_i) + \|a\|^2_{\ell_2} - R.
\]
If $f(x)>0$, then a pattern $x$ is located outside the sphere and is considered to be anomalous. 

\section{Privileged Information}

Let us assume that during the training phase we have some privileged information: besides patterns $(x_1,\ldots,x_l)\in\mathcal{X}\subset\RR^n$ (original information) we also have additional patterns $(x_1^*,\ldots,x_l^*)\in\mathcal{X}^*\subset\RR^m$. This additional (privileged) information is not available on the test phase, i.e. we are going to train our decision rule on pairs of patterns $(x, x^*)\in\mathcal{X}\cup\mathcal{X}^*\subset\RR^{n+m}$, but when making decisions we can use only test patterns $x\in\mathcal{X}$.

Let us discuss how this privileged information can be incorporated in the considered problem statements. In the original approaches to one-class classification we assume that the slack variables $\xi_i$, characterizing the distance from the patterns $x_i$ to the separating boundary, are determined through the solution of the corresponding optimization problem (see \eqref{oneclass} or \eqref{svddopt}). Now let us assume that the slack variables can be modelled as 
\begin{equation}
\label{slackmod}
\xi_i = \xi_i(x^*_i) = (\phi^*(x^*_i)\cdot w^*) + b^*,
\end{equation}
where $\phi^*(\cdot)$ is a feature map in the space of privileged patterns. Thus, we assume that using the privileged patterns $(x_1^*,\ldots,x_l^*)$ we can refine the location of the separating boundary w.r.t. the sample of training objects. 

\subsection{One-Class SVM+}
Let us modify problem statement \eqref{oneclass} in order to incorporate privileged information:
\begin{eqnarray}\label{ocsvm+} 
\frac{\nu l}{2}\|w\|^2_{\ell_2} + \frac{\gamma}{2}\|w^*\|^2_{\ell_2} - \nu l\rho + \notag\\
\sum\limits_{i=1}^l\left[(w^*\cdot \phi^*(x_i^*)) + b^* + \zeta_i\right]\to\min\limits_{w, w^*, b^*, \rho, \zeta} \\
s.t.\,\,\,\,(w\cdot \phi(x_i)) \geq \rho - (w^*\cdot\phi^*(x^*_i)) - b^*,\notag\\
(w^*\cdot\phi^*(x^*_i)) + b^* + \zeta_i \geq 0,\,\,\,\,\zeta_i \geq 0.\notag
\end{eqnarray}
Here $\gamma$ is a regularization parameter for the linear approximation of the slack variables, $\zeta_i$ are instrumental variables used to prevent those patterns, belonging to a ``positive'' half-plane, from being penalized. Note that if $\gamma\to\infty$, then the solution of \eqref{ocsvm+} is close to the original solution of \eqref{oneclass}.

Let us write out a Lagrangian for \eqref{ocsvm+}:
\begin{eqnarray}
L = \frac{\nu l}{2}\|w\|^2_{\ell_2} - \nu l \rho + \frac{\gamma}{2}\|w^*\|^2_{\ell_2} \nonumber  \notag\\
+\sum\limits_{i=1}^l\left[(w^*\cdot \phi^*(x_i^*)) + b^* + \zeta_i\right]- \sum\limits_{i=1}^l\mu_i\zeta_i \nonumber \\ 
 - \sum\limits_{i=1}^l\alpha_i\left[(w\cdot \phi(x_i)) - \rho +(w^*\cdot \phi^*(x^*_i)) + b^*\right] \notag\\
 -\sum\limits_{i=1}^l\beta_i\left[(w^*\cdot \phi^*(x_i^*)) + b^* + \zeta_i\right]. \notag
\end{eqnarray}
Setting $\delta_i = 1 - \beta_i$, from the Karush -- Kuhn -- Tucker conditions we get that
\begin{eqnarray}
w = \frac{1}{\nu l}\sum\limits_{i=1}^l\alpha_i\phi(x_i),\,\,\,\, w^* = \frac{1}{\gamma}\sum\limits_{i=1}^l(\alpha_i - \delta_i) \phi^*(x_i^*),\notag\\
\delta_i = \mu_i,\,\,\,\, \sum\limits_{i=1}^l\delta_i = \sum\limits_{i=1}^l\alpha_i = \nu l,\,
0 \leq \delta_i \leq 1.\notag
\end{eqnarray}

Using obtained equations, we now can formulate the dual problem
\begin{eqnarray}
-\frac{1}{2\nu l}\sum\limits_{i,j}\alpha_i\alpha_jK(x_i, x_j) \notag\\
- \sum\limits_{i,j}\frac{1}{2\gamma}(\alpha_i - \delta_i)K^*(x_i^*, x_j^*)(\alpha_j - \delta_j)\to\max\limits_{\alpha, \delta}\notag\\
s.t.\,\,\,\sum\limits_{i=1}^l\alpha_i = \nu l,\,\,\,
\sum\limits_{i=1}^l\delta_i = \nu l,\,\,\, 0 \leq \delta_i \leq 1,\,\,\, \alpha_i \geq 0\notag.
\end{eqnarray}
Here we replace the scalar product $(\phi^*(x^*_i)\cdot \phi^*(x^*_j))$ with the corresponding kernel function $K^*(x^*_i, x^*_j)$. At the end, the decision function has the same form as in the case of the original One-Class SVM: $f(x) = \sum\limits_{i=1}^l \alpha_i K(x_i, x) - \rho.$

\subsection{SVDD+}

Let us modify problem statement \eqref{svddopt} in order to incorporate privileged information:
\begin{eqnarray}\label{svddopt+}
\nu l R + \frac{\gamma}{2}\|w^*\|^2_{\ell_2} \notag\\
+ \sum\limits_{i=1}^l\left[(w^*\cdot \phi^*(x_i^*)) + b^* + \zeta_i\right]\to\min\limits_{R,a, w^*, b, \zeta}\,\,\,\,\,\, \\
s.t.\,\|\phi(x_i) - a \|^2_{\ell_2} \leq R 
+ [(w\cdot \phi^*(x_i^*)) + b^*],\notag\\
(w^*\cdot \phi^*(x_i^*)) + b^* + \zeta_i \geq 0,\,\zeta_i \geq 0. \notag
\end{eqnarray}
If $\gamma\to\infty$, then the solution of \eqref{svddopt+} is close to the original solution of \eqref{svddopt}.

Let us write out a Lagrangian for \eqref{svddopt+}:
\begin{eqnarray}
L = \nu l R + \frac{\gamma}{2}\|w^*\|^2_{\ell_2} + \sum\limits_{i=1}^l[(w^*\cdot \phi^*(x_i^*)) + b^* + \zeta_i]- \sum\limits_{i=1}^l \mu_i\zeta_i  \nonumber\\
+\sum\limits_{i=1}^l\alpha_i[\|\phi(x_i) - a\|^2_{\ell_2} - R - (w^*\cdot \phi^*(x_i^*)) -b^*]\notag\\
-\sum\limits_{i=1}^l\beta_i[(w^*\cdot \phi^*(x_i^*)) + b^* + \zeta_i].\notag
\end{eqnarray}
Setting $\delta_i = 1 - \beta_i$, from the Karush -- Kuhn -- Tucker conditions we get that
\begin{eqnarray}
w^* = \frac{1}{\gamma}\sum\limits_{i=1}^l(\alpha_i - \delta_i) \phi^*(x_i^*),\,
a=\frac{1}{\nu l}\sum\limits_{i=1}^l\alpha_i\phi(x_i),\notag\\
\delta_i = \mu_i,\,\sum\limits_{i=1}^l\alpha_i = \sum\limits_{i=1}^l\delta_i = \nu l,\,
0 \leq \delta_i \leq 1.\notag
\end{eqnarray}
Let us formulate the dual problem:
\begin{eqnarray}
\sum\limits_{i=1}^l \alpha_i K(x_i, x_i) - \frac{1}{2\nu l}\sum\limits_{i, j} \alpha_i\alpha_jK(x_i, x_j) \notag\\
-\sum\limits_{i, j}\frac{1}{2\gamma}(\alpha_i - \delta_i)K^*(x_i^*, x_j^*)(\alpha_j - \delta_j)\to\max\limits_{\alpha, \delta}\notag\\
s.t.\,\,\,
\sum\limits_{i=1}^l\alpha_i = \nu l,\,\,\sum\limits_{i=1}^l\delta_i = \nu l,\,\, 0\leq \delta_i \leq 1,\,\, \alpha_i \geq 0.\notag
\end{eqnarray}
At the end, the decision function has the same form as in the case of the original SVDD:
$f(x) = K(x, x) - 2\sum\limits_{i=1}^l\alpha_i K(x, x_i) + \|a\|^2_{\ell_2} - R.$

\section{Related work}
\label{sec:otherTechniques}

In principle some attempts to use privileged information for one-class classification can be found in the literature. E.g. in \cite{ocsvmother} a feature space is divided into a number of meaningful subdomains $\mathcal{X}=\cup_{r=1}^N\mathcal{X}_r$ and for each element from the training sample $(x_1,\ldots, x_l)$ ordinal number of the subdomain, to which this element belongs to, is used as privileged information in order to introduce different constraints on the slack variables for patterns from different subdomains, i.e. $\xi_{i,r} = \xi_{r}(x_i) = (\phi^*_r(x_i)\cdot w^*_r) + b^*_r,$ $x_i\in\mathcal{X}_r$, $r = 1,\ldots,N$.

Thus the authors do not explicitly use privileged information from some other feature space rather then initial one: privileged information can be straightforwardly calculated from values of the original patterns $(x_1,\ldots,x_l)$.

In order to construct an anomaly detection rule the authors propose to solve the following optimization problem:
\begin{eqnarray}\label{eq:one_class_svmnewnew}
\frac{1}{2}\|w\|^2_{\ell_2} + \frac{\gamma}{2}\sum_{r=1}^N\|w^*_r\|^2_{\ell_2} + \nonumber \\ 
\frac{1}{N}\sum_{r=1}^N\frac{1}{\nu_r}\sum_{x_i\in\mathcal{X}_r}[(w^*_r\cdot \phi^*_r(x_i)) + b^*_r] -\rho\to \min\limits_{w, w^*_r, b^*_r, \rho, \zeta_r}\\
s.t.\,\,\,(w\cdot \phi(x_i)) \geq \rho - (w^*_r\cdot \phi^*_r(x_i)) - b^*_r,x_i\in\mathcal{X}_r\nonumber,\\
(w^*_r\cdot \phi^*_r(x_i)) + b^*_r + \zeta_{i,r} \geq 0,x_i\in\mathcal{X}_r\nonumber,\\
\zeta_{i,r} \geq 0\nonumber.
\end{eqnarray}
A shortcoming of the problem statement  \eqref{eq:one_class_svmnewnew} (cf. with the problem statement \eqref{ocsvm+}) is that the parameters $\nu_r$ and $\gamma$ influence the regularization in a dependent manner, i.e. their contribution to the regularization can not be disentangled.

A similar framework is used in \cite{svddother}, where, as in the previous paper, the authors use ordinal numbers of corresponding subdomains as privileged information. The main difference with paper  \cite{ocsvmother} is that the SVDD algorithm underlies their approach. In fact the authors of \cite{svddother} propose to solve the following optimization problem in order to find the decision rule:

\begin{eqnarray}\label{eq:svddnewnew}
R^2 + \frac{\gamma}{2}(R^{*})^2 \nonumber\\
+ \frac{1}{\nu l}\sum_{i=1}^l[\|\phi^*(x_i^*) - a^*\|_{\ell_2}^2 - (R^{*})^2] \to \min\limits_{R, R^*, a, a^* ,\zeta}\\
s.t.\,\,\,\|\phi(x_i) - a\|_{\ell_2}^2 \leq R^2+\|\phi^*(x_i^*) - a^*\|_{\ell_2}^2 - (R^{*})^2, \nonumber\\
\|\phi^*(x_i^*) - a^*\|_{\ell_2}^2 - (R^{*})^2 + \zeta_i \geq 0\nonumber,\\
\zeta_i \geq 0\nonumber.
\end{eqnarray}

As in the previous example, the parameterization, used in (\ref{eq:svddnewnew}), does not allow to control the regularization in the original feature space and in the space of privileged information independently. One more difference from the problem statement, proposed in this paper, is another approach to modelling slack variables $\xi_i$. In our approach we use linear model \eqref{slackmod}, but in \cite{svddother} the distance $\|\phi^*(x_i^*) - a^*\|_{\ell_2}^2 - (R^{*})^2$ to the surface of the sphere in the privileged space is used.

\section{Numerical experiments}
In this section we describe performed numerical experiments. We provide results only for the original One-Class Support Vector Machine and its elaborated extension with privileged information, since results, based on SVDD, are comparable. This is not surprising, since when the Gaussian kernel is used the decision rules for the both methods are essentially the same, as it follows from the theoretical results of \cite{svddr}.

\subsection{Data description}
\label{subsec:data_desc}
We perform experiments using both two-dimensional synthetic data and some real data from the Microsoft Malware Classification Challenge (BIG 2015), see \cite{malwa}.

We generate the first synthetic dataset (``Mixture of Gaussians'') from a mixture of two-dimensional normal distributions, with mean values $c_1 = (2, 2)$ and $c_2 = (-2, -2)$ and unit covariance matrices ${I}_2\in\mathbb{R}^{2\times2}$, i.e. 
\[
x\sim \eta\cdot\mathcal{N}(c_1,{I}_2) + (1-\eta)\cdot\mathcal{N}(c_2,{I}_2),\,\,\eta\sim\mathrm{Ber}(0.5).
\] 
As privileged information we use coordinates of a pattern after subtracting the nearest mean vector, i.e. $x^* = x - \arg\min_{c_1, c_2} (\|x-c_1\|, \|x-c_2\|)$.

We represent the second synthetic dataset (``Circles'') by two circles of different radii $r_1$ and $r_2$ with common center. We generate each circle in polar coordinate system, then Cartesian coordinates of a pattern are calculated: 
\begin{align*}
x &= (r\cdot\cos\phi,\,r\cdot\sin\phi),\\
r &= \eta\cdot\mathrm{I}(\eta>0),\,\eta\sim\mathcal{N}(r_0,0.5),\\
\phi&\sim U[0,2\pi),
\end{align*}
where $\mathrm{I}(\cdot)$ is an indicator function, $U[0, 2\pi)$ is a uniform distribution, $r_0 = 5$ for the external circle and $r_0=0.5$ for the internal circle. We use Cartesian coordinates as features and polar coordinates $(r,\phi)$ as  privileged information.

We generate the third synthetic dataset (``Arc'') in polar coordinates: 
\begin{align*}
\phi&\sim\mathcal{N}(0,0.04),\\
\tau& = \eta\cdot(0.1-|\phi|),\,\eta\sim\mathcal{N}(-1/2,1),\\
x &= ((10-\tau)\cdot\cos\phi,\,(10-\tau)\cdot\sin\phi),\\
\end{align*}
where thanks to the multiplier $0.1-|\phi|$ we get bigger variance for $\phi\approx0$ and as a consequence a banana-like shape of the dataset. We use Cartesian coordinates as features and polar coordinates as  privileged information.

For some experiments we need to generate a sample with noise:
\begin{itemize}
\item generate a sample without noise, 
\item estimate bounds $[x_{1,\min},x_{1,\max}]\times[x_{2,\min},x_{2,\max}]$ of the sample,
\item generate noise using uniform distribution $U[a_1, a_2] \times U[b_1, b_2]$, where $a_1=x_{1,\min} - 0.5\cdot(x_{1,\max} - x_{1,\min})$, $a_2 = x_{1,\max} + 0.5\cdot(x_{1,\max} - x_{1,\min})$, $b_1 = x_{2,\min} - 0.5\cdot(x_{2,\max} - x_{2,\min})$, $b_2 = x_{2,\max} + 0.5\cdot(x_{2,\max} - x_{2,\min})$. 
\end{itemize}
Thus the region of the feature space, in which anomalies are located, includes all patterns of a ``normal'' data.

\begin{figure}[t!]
\begin{subfigure}{.25\textwidth}
  \centering
  \includegraphics[width=.99\linewidth]{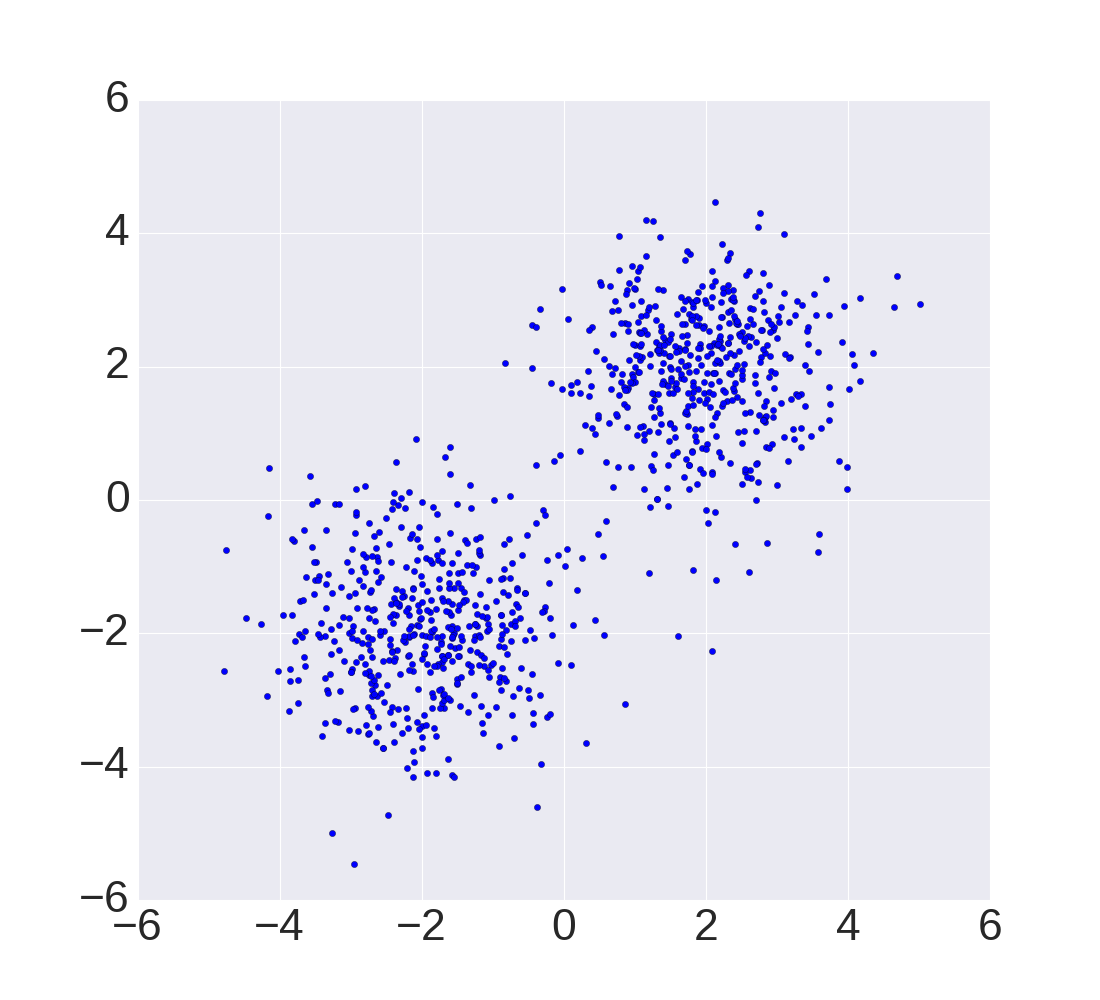}
  \caption{Mixture of Gaussians\\\hspace{\textwidth}}
\end{subfigure}%
\begin{subfigure}{.25\textwidth}
  \centering
  \includegraphics[width=.99\linewidth]{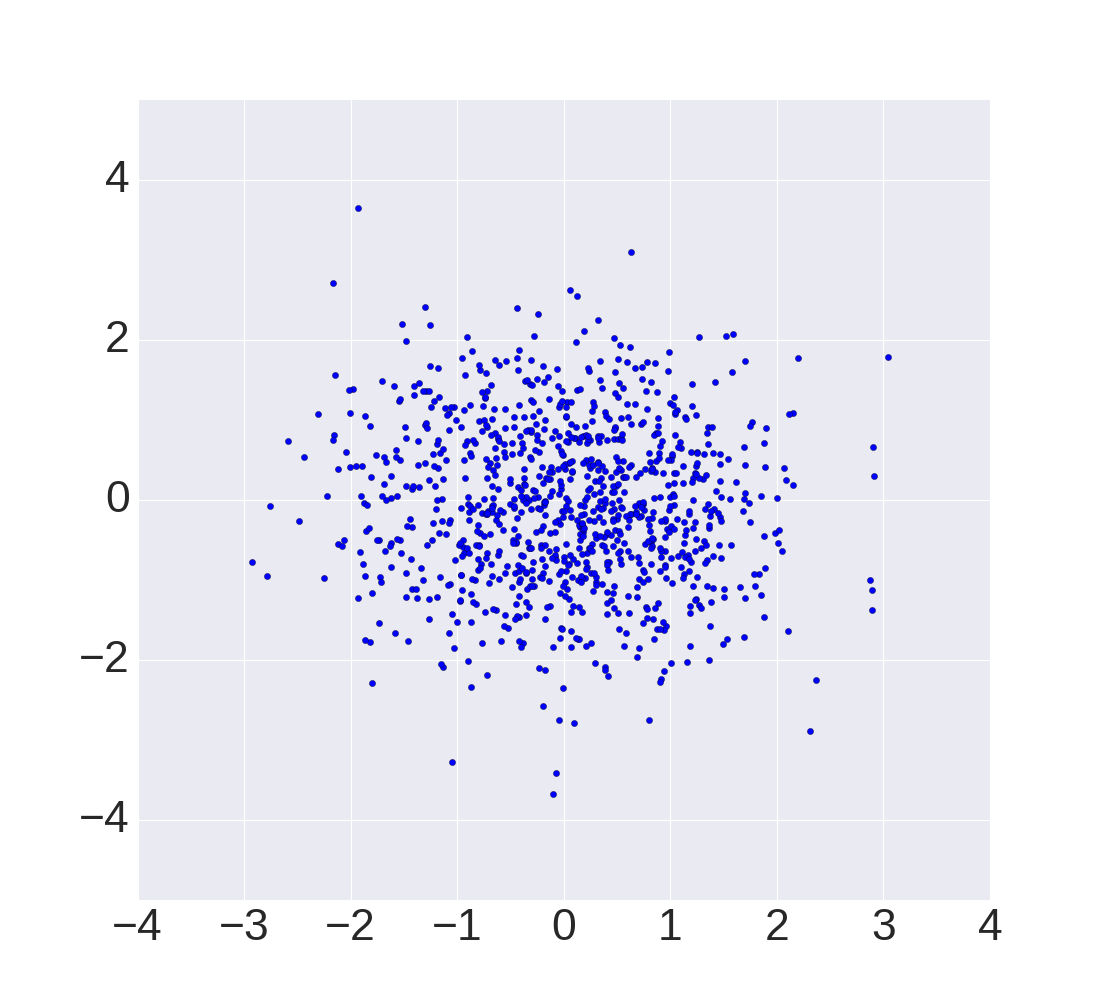}
  \caption{Mixture of Gaussians, privileged feature space}
\end{subfigure}
\begin{subfigure}{.25\textwidth}
  \centering
  \includegraphics[width=.99\linewidth]{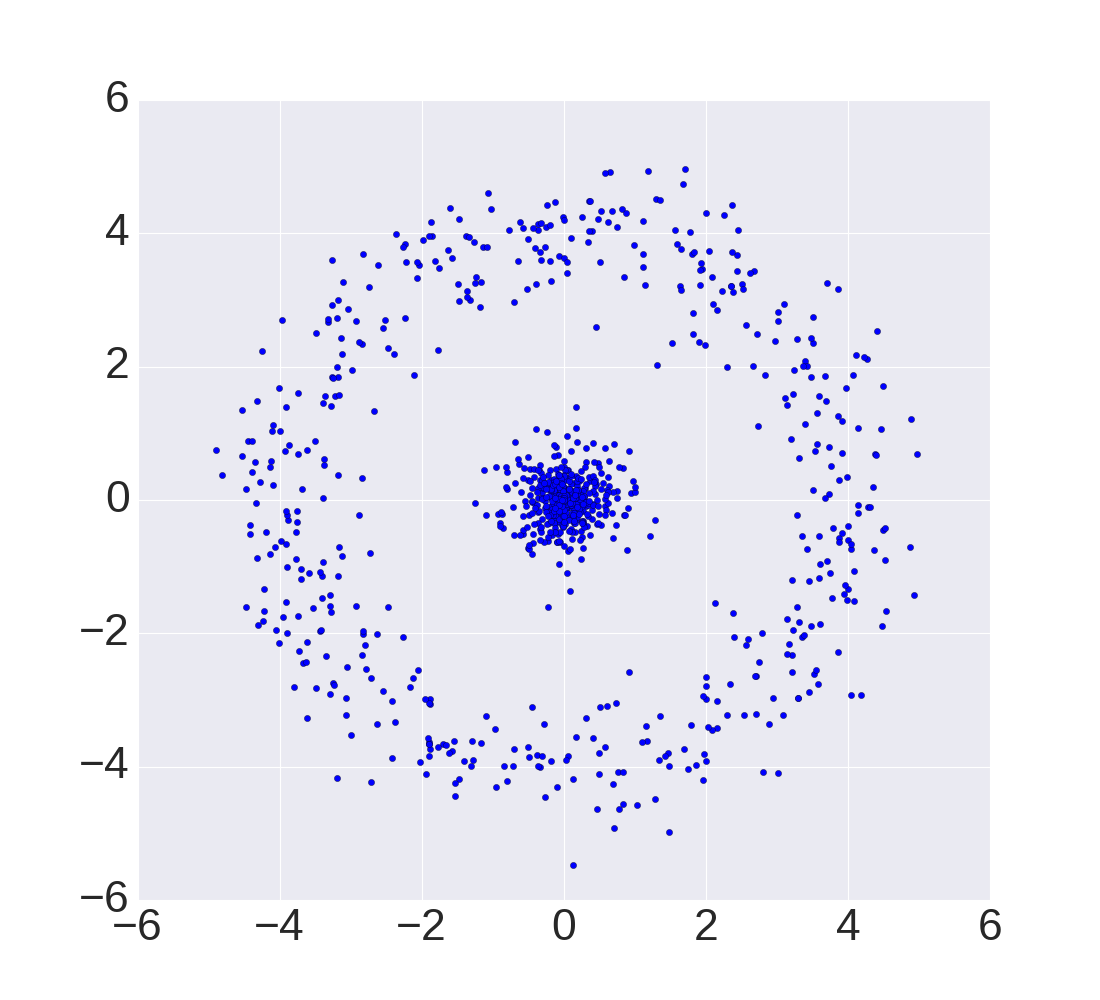}
  \caption{Circles\\\hspace{\textwidth}}
\end{subfigure}%
\begin{subfigure}{.25\textwidth}
  \centering
  \includegraphics[width=.99\linewidth]{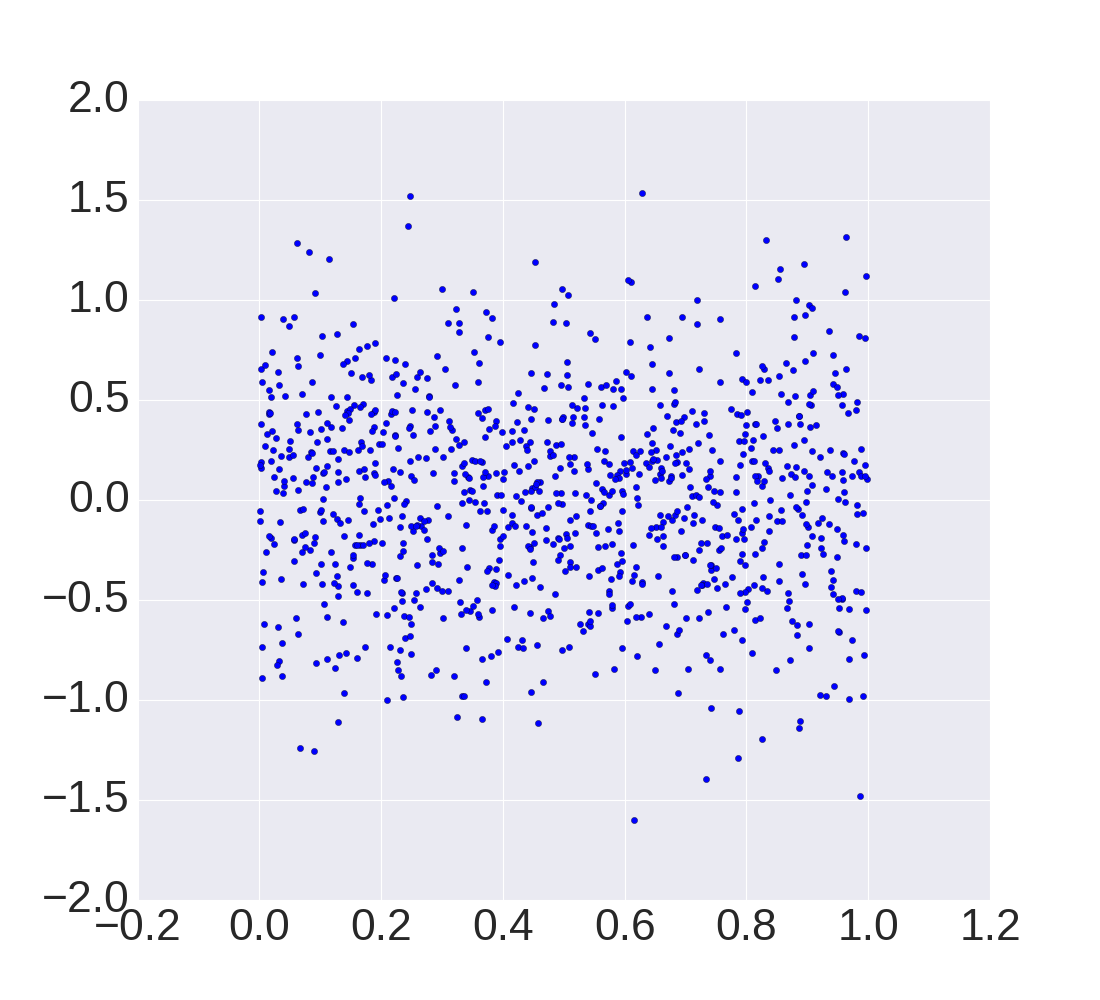}
  \caption{Circles, privileged feature space}
\end{subfigure}
\begin{center}
\begin{subfigure}{.25\textwidth}
  \centering
  \includegraphics[width=.99\linewidth]{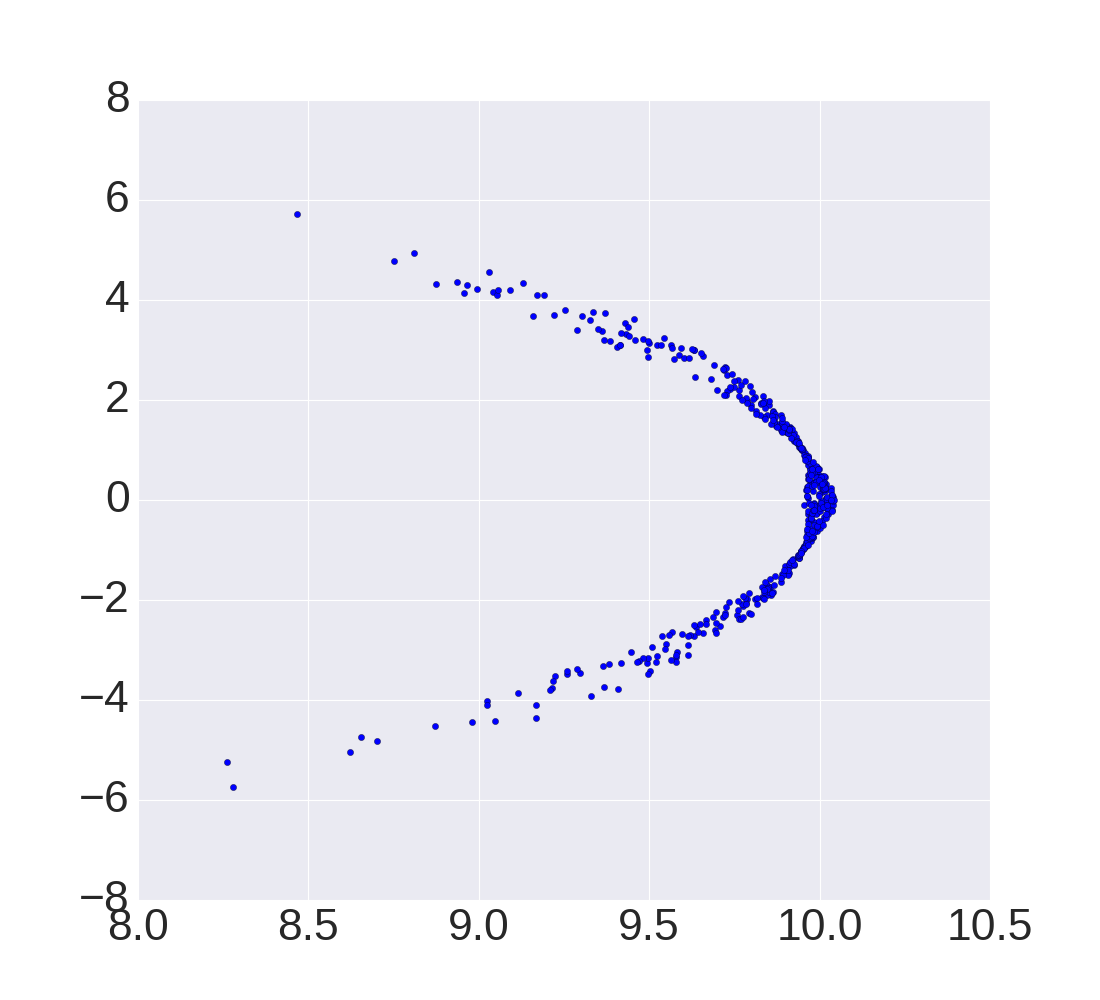}
  \caption{Arc\\\hspace{\textwidth}}
\end{subfigure}%
\begin{subfigure}{.25\textwidth}
  \centering
  \includegraphics[width=.99\linewidth]{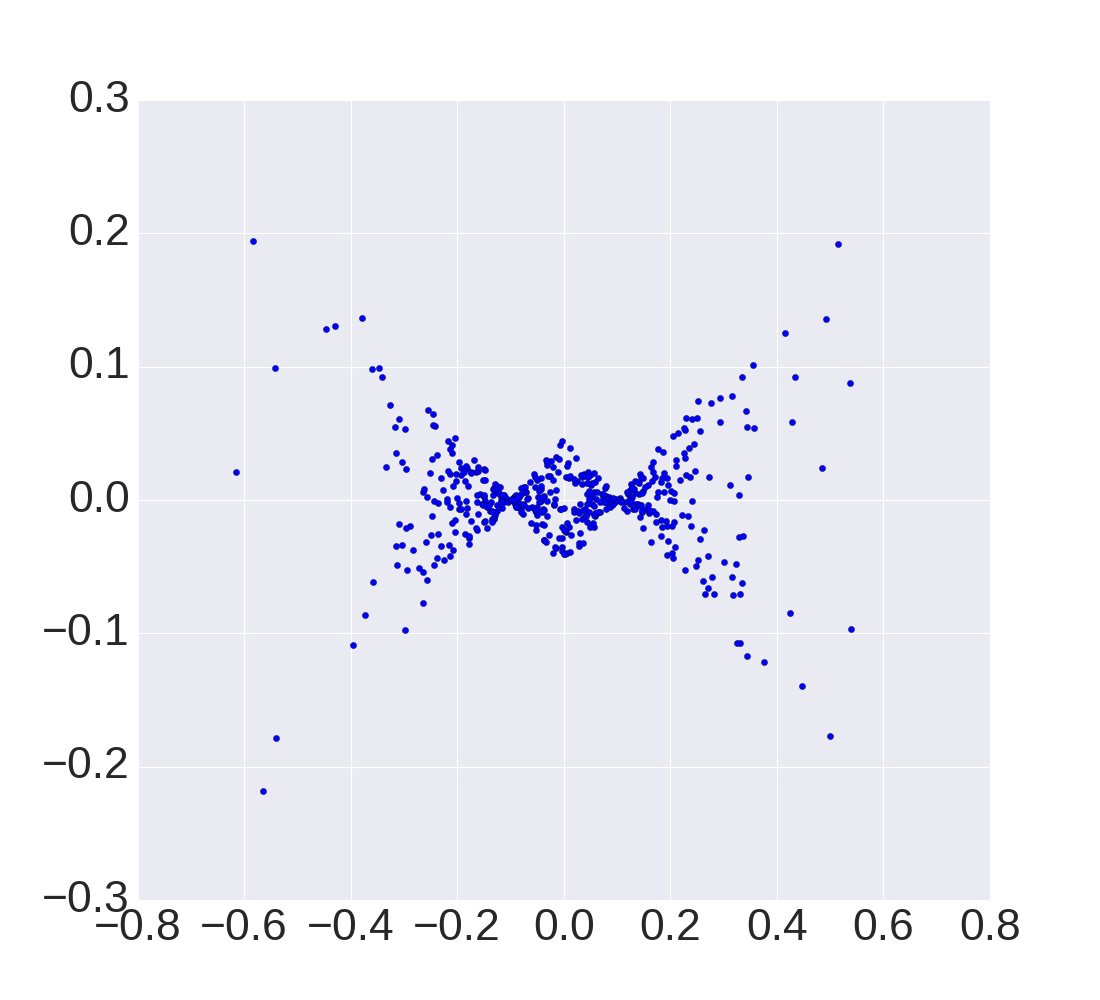}
  \caption{Arc, privileged feature space}
\end{subfigure}%
\end{center}
\caption{Examples of synthetic datasets}
\label{examples}
\end{figure}

Examples of synthetic data are given in figure \ref{examples}.

\subsection{Proportion of discarded test sample patterns}

In case of the original One-Class SVM a number of test patterns, marked by the decision rule as anomalous, depends on the regularization parameter $\nu$ \cite{nusvm,svdd} in a very certain way: it tends to $\nu$ if the training sample size increases and both the test sample and the train sample are generated from the same distribution. 

\begin{figure}[t!]
\begin{center}
\begin{subfigure}{.25\textwidth}
  \centering
  \includegraphics[width=.99\linewidth]{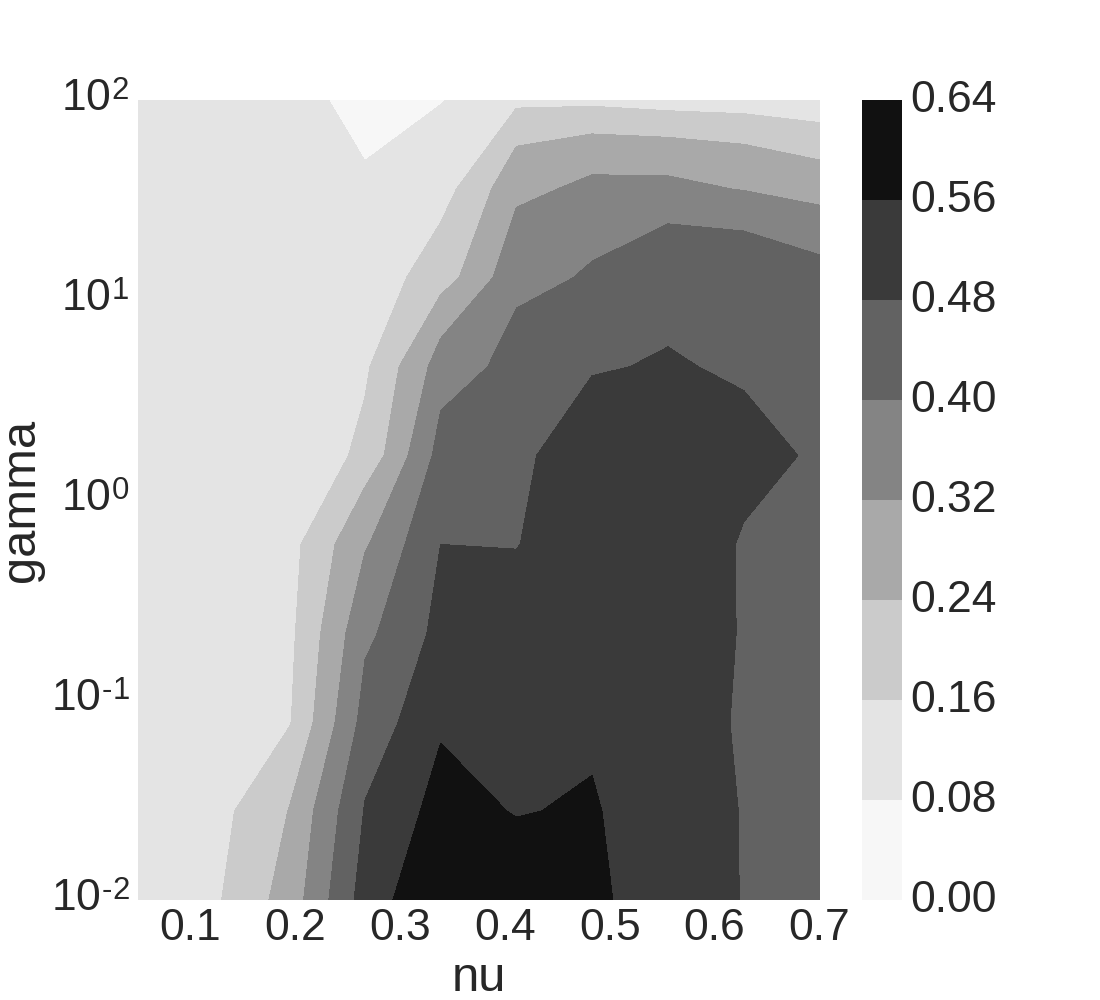}
  \caption{Arc}
\end{subfigure}%
\begin{subfigure}{.25\textwidth}
  \centering
  \includegraphics[width=.99\linewidth]{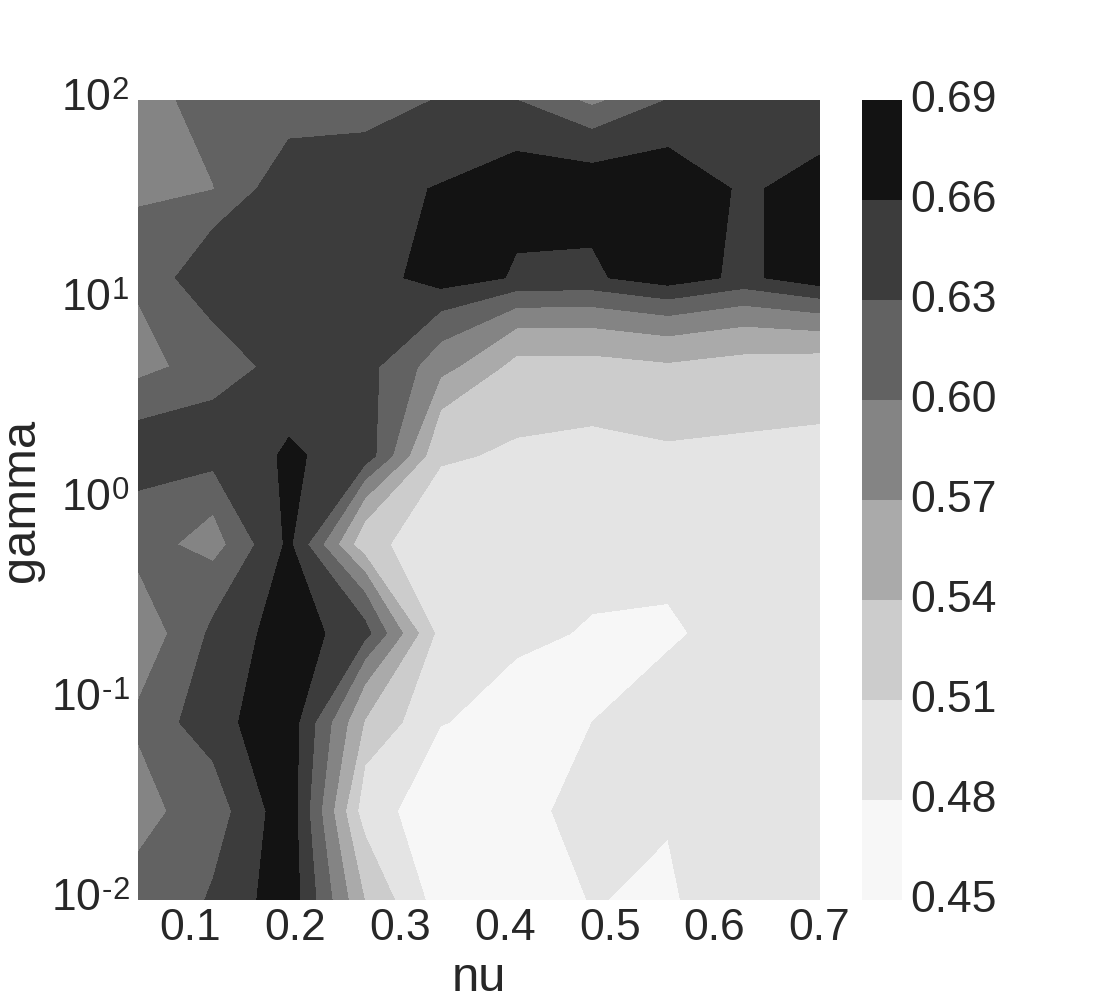}
  \caption{Circles}
\end{subfigure} %
\begin{subfigure}{.25\textwidth}
  \centering
  \includegraphics[width=.99\linewidth]{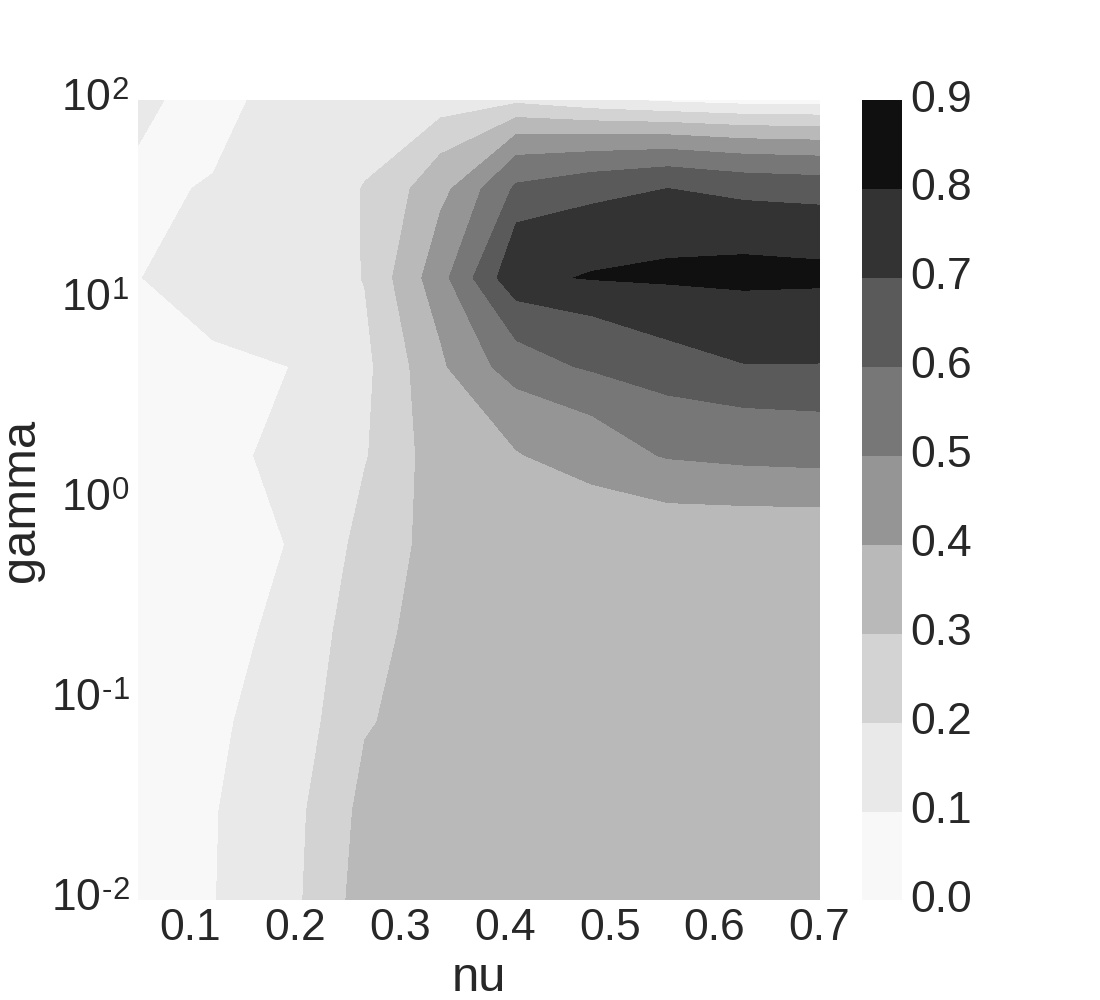}
  \caption{Mixture of Gaussians}
\end{subfigure}%
\caption{Synthetic datasets. Proportion of test sample patterns, marked as anomalous by One-Class SVM+}
\label{clean}
\end{center}
\end{figure}

Using synthetic datasets let us check how the proportion of test sample patterns, marked as anomalous, depends on the regularization parameters $\nu$ and $\gamma$. We use the Gaussian kernel $K(x, x') = \exp(-\|x - x'\|^2/\sigma^2)$ with $\sigma^2 = 2$ both for the original feature space, and for the privileged feature space. Results are provided in figure \ref{clean}. We can notice that for small values of $\nu$ (significant regularization of the privileged space) this dependence is similar with that for the original One-Class SVM.

\subsection{Accuracy of Anomaly Detection}

\begin{figure}[t!]
\begin{center}
\begin{subfigure}{.25\textwidth}
  \centering
  \includegraphics[width=.99\linewidth]{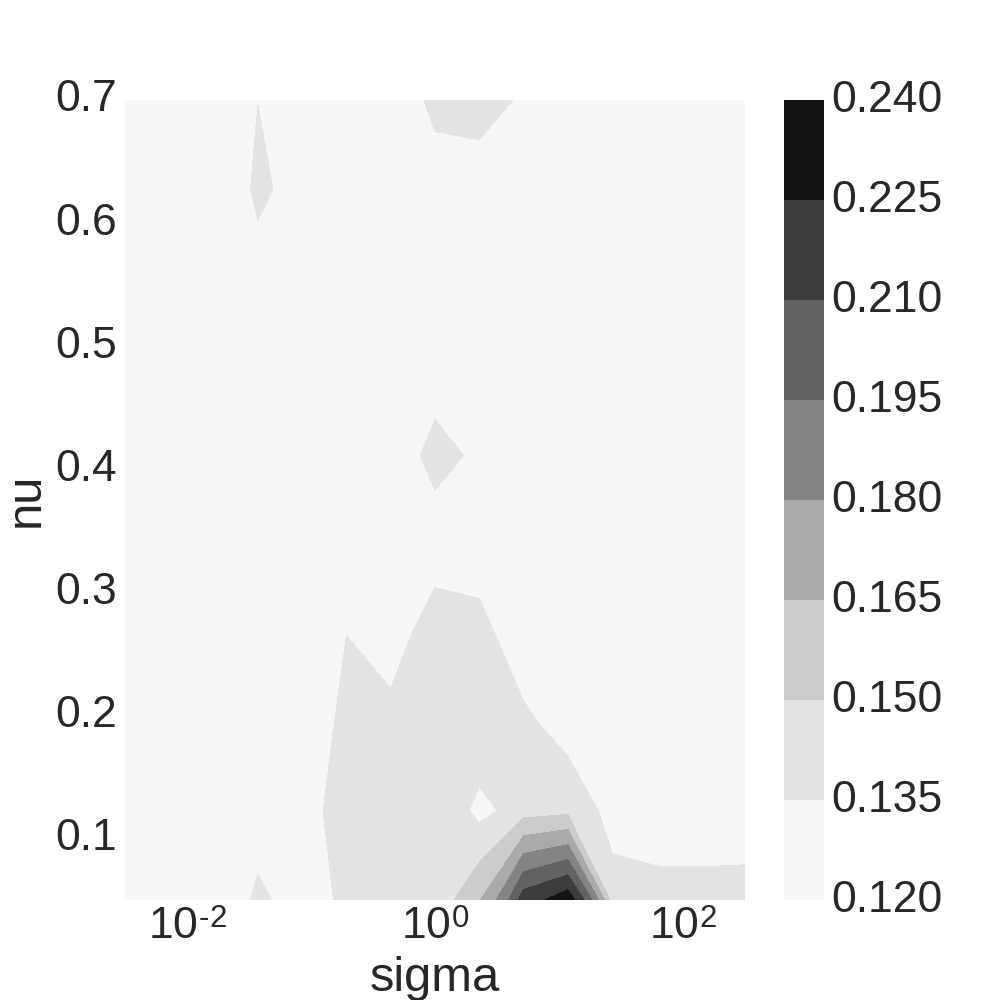}
  \caption{Arc}
\end{subfigure}%
\begin{subfigure}{.25\textwidth}
  \centering
  \includegraphics[width=.99\linewidth]{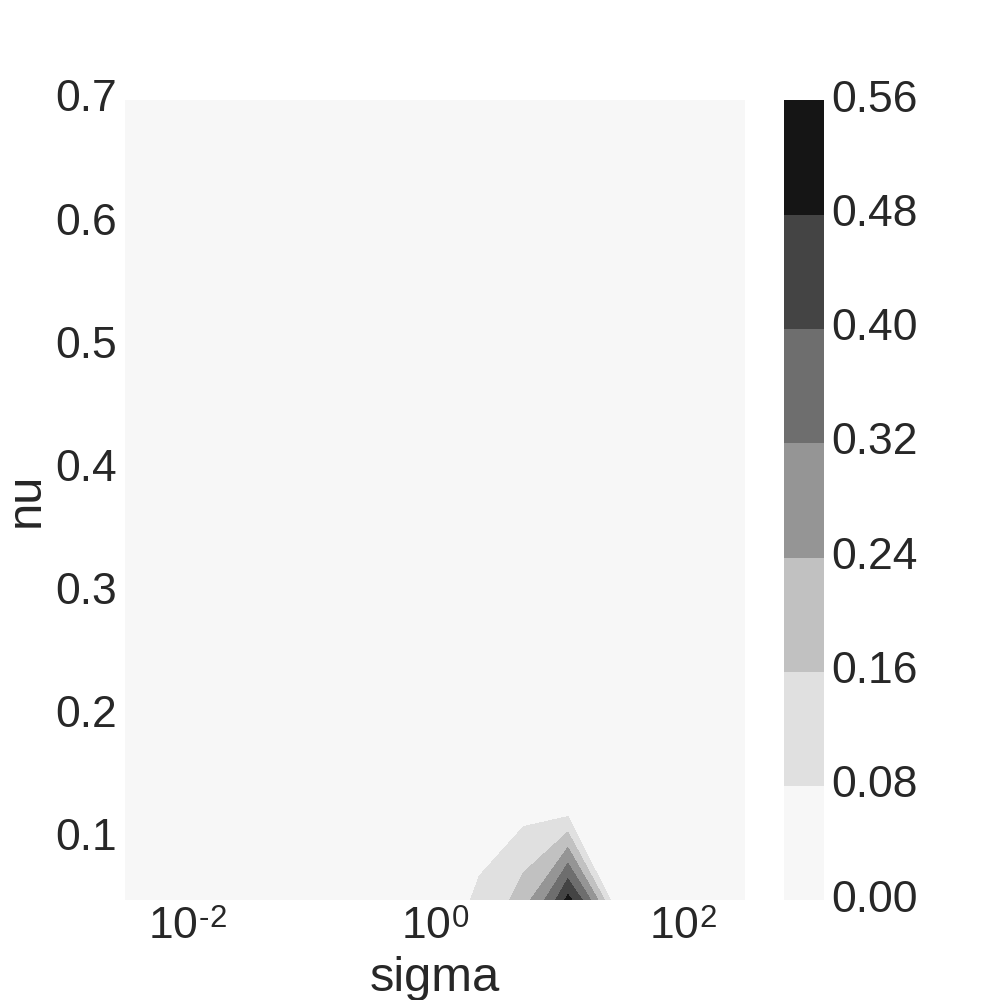}
  \caption{Circles}
\end{subfigure} %
\begin{subfigure}{.25\textwidth}
  \centering
  \includegraphics[width=.99\linewidth]{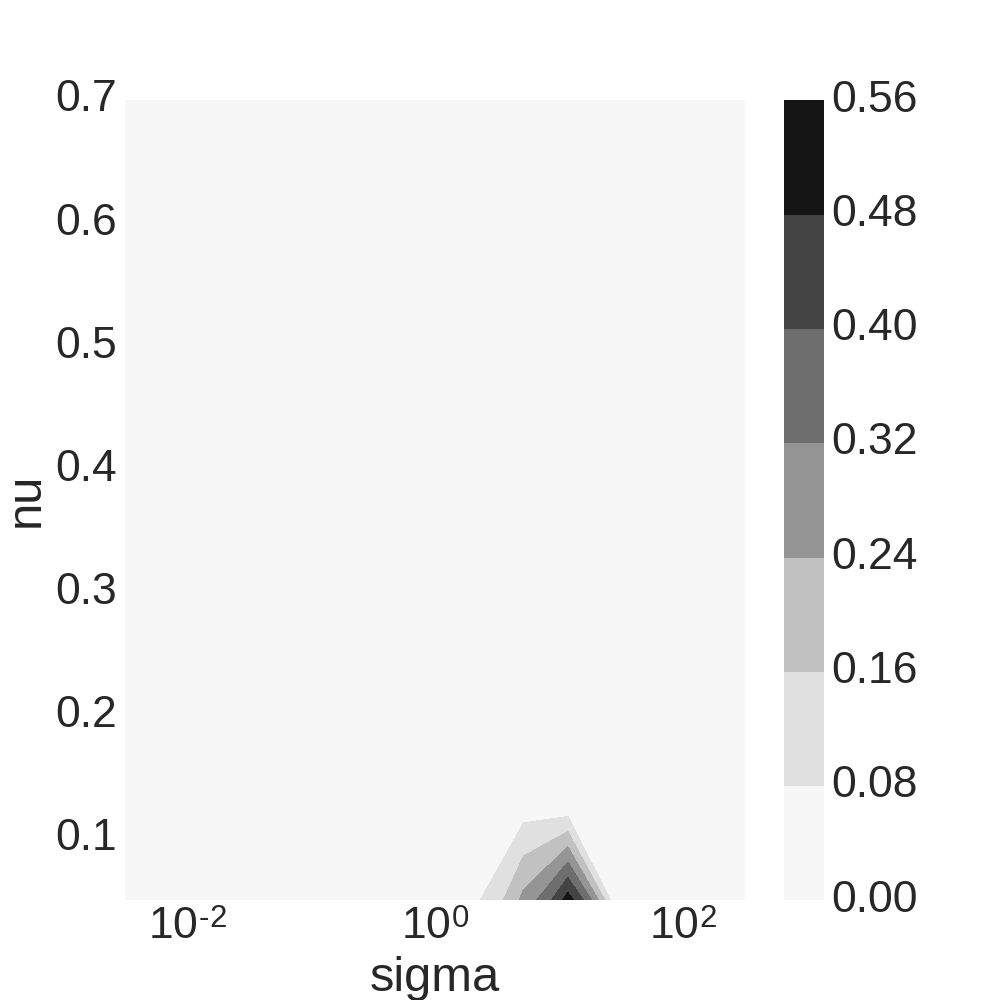}
  \caption{Mixture of Gaussians}
\end{subfigure}%
\caption{Synthetic datasets. Area under the precision/recall curve for the original One-Class SVM}
\label{noisy_ordinary}
\end{center}
\end{figure}

\begin{figure}[t!]
\begin{center}
\begin{subfigure}{.25\textwidth}
  \centering
  \includegraphics[width=.99\linewidth]{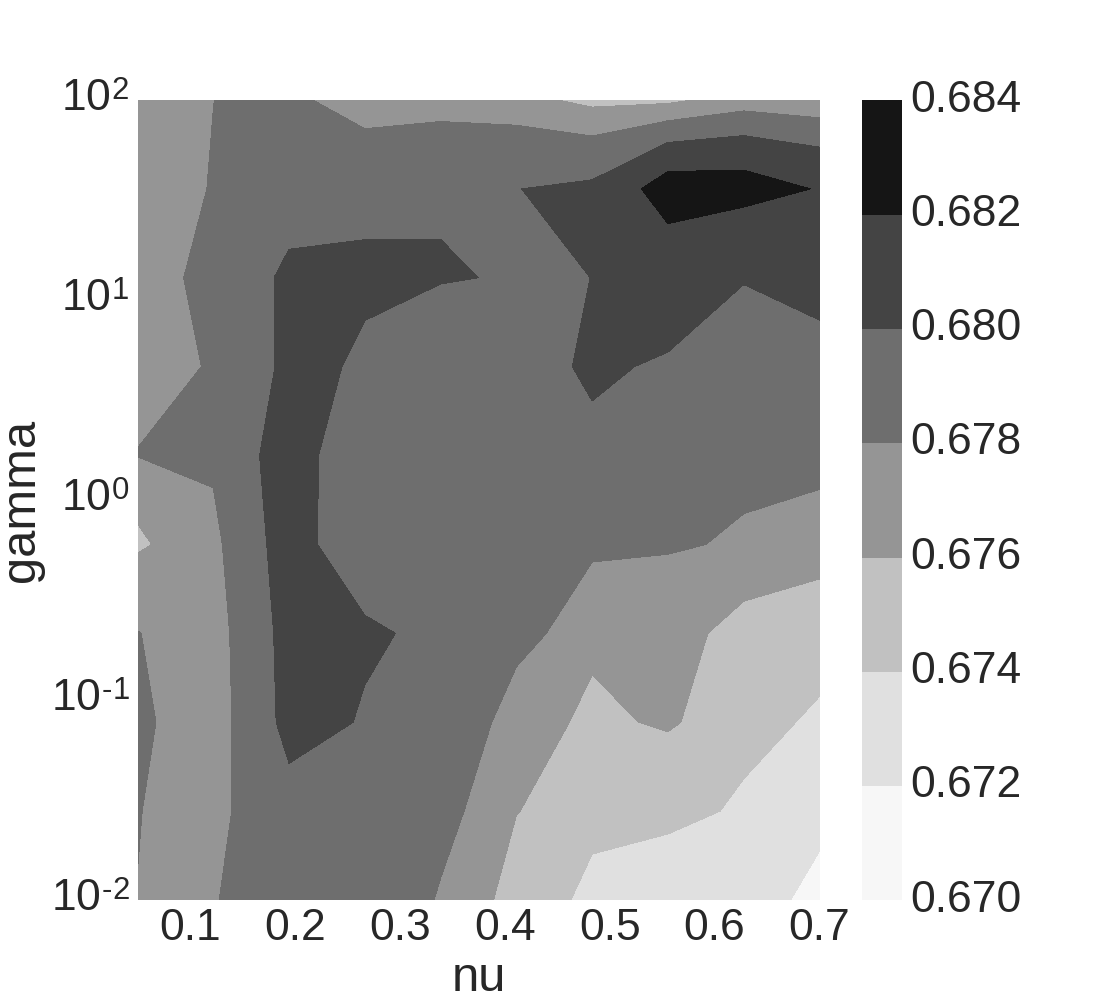}
  \caption{Arc}
\end{subfigure}%
\begin{subfigure}{.25\textwidth}
  \centering
  \includegraphics[width=.99\linewidth]{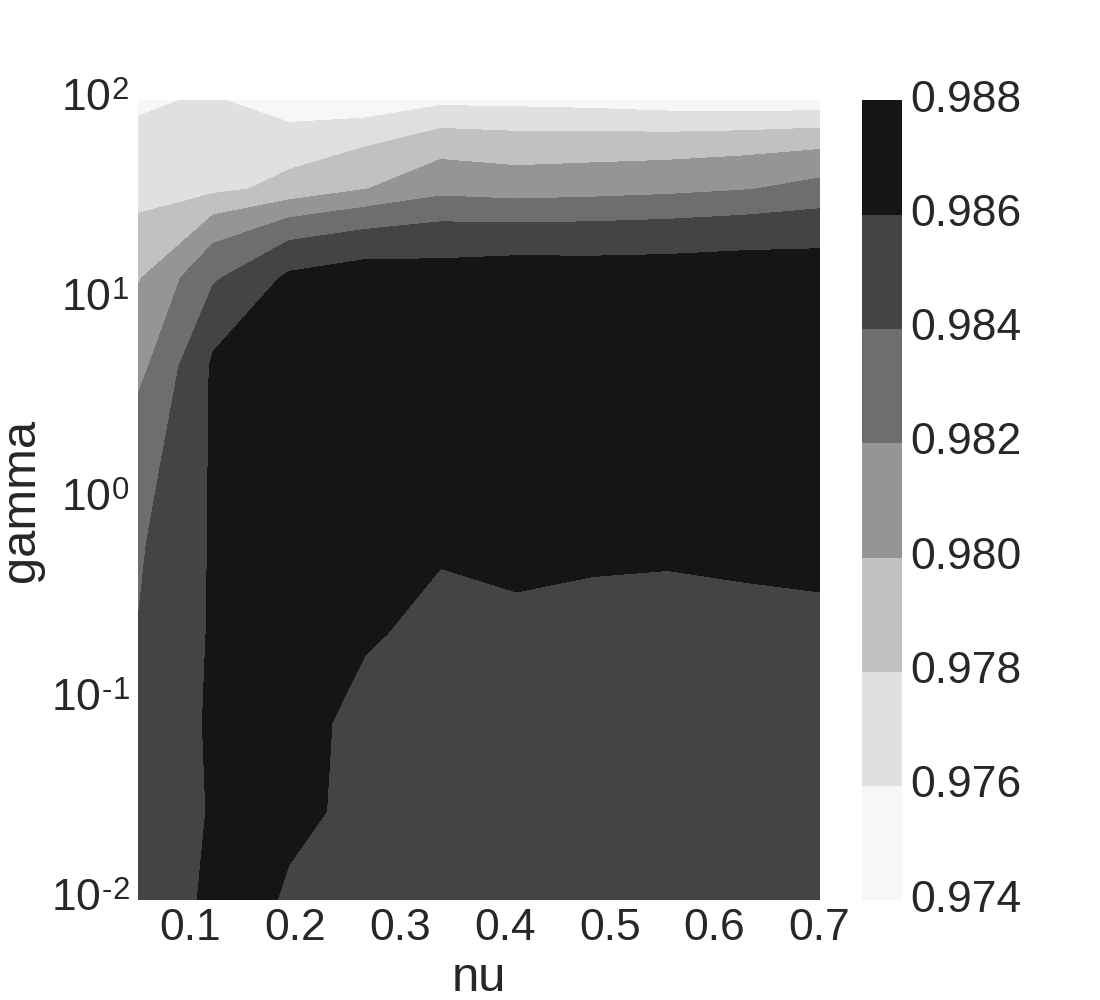}
  \caption{Circles}
\end{subfigure}
\begin{subfigure}{.25\textwidth}
  \centering
  \includegraphics[width=.99\linewidth]{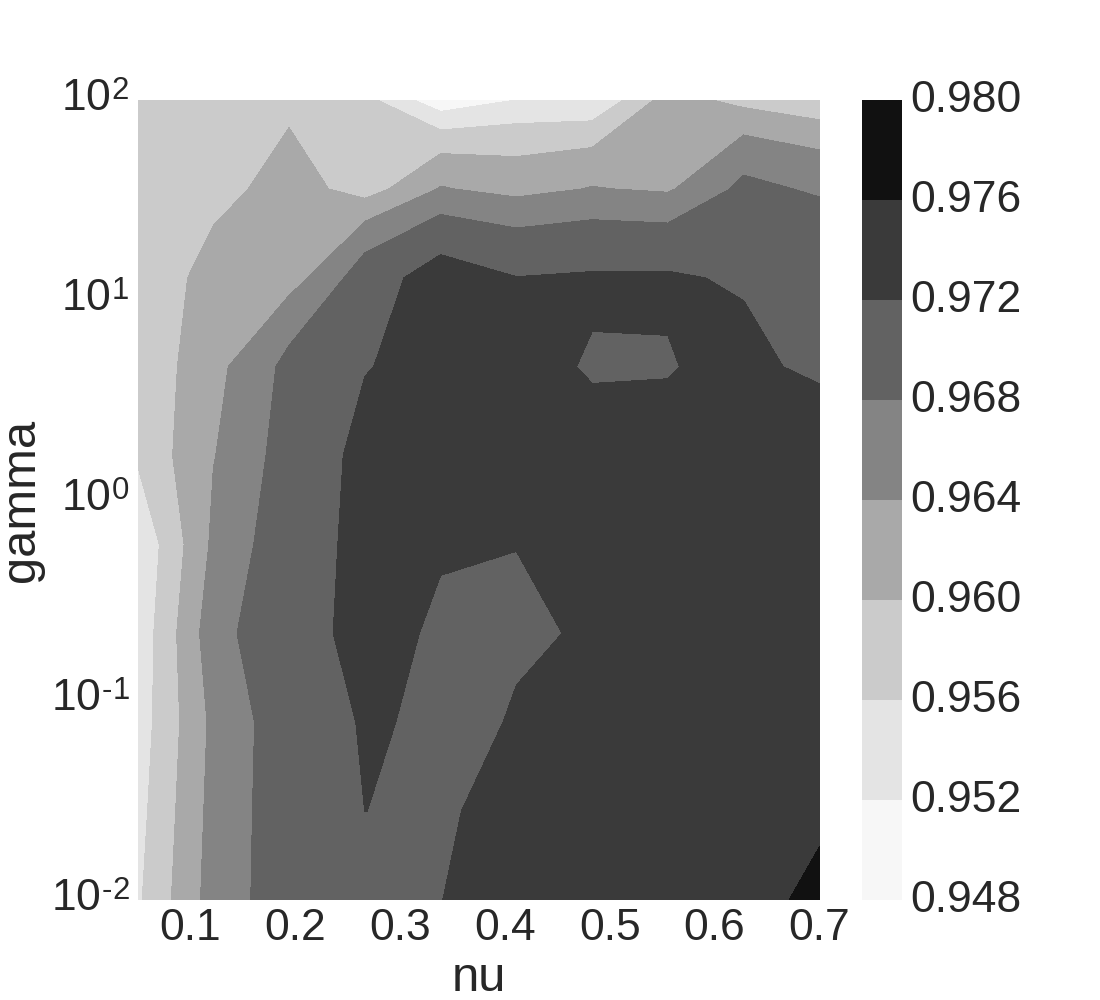}
  \caption{Mixture of Gaussians}
\end{subfigure}%
\caption{Synthetic datasets. Area under the precision/recall curve for the One-Class SVM+}
\label{noisy}
\end{center}
\end{figure}

For this experiment we generate samples with outliers using the approach, described in subsection \ref{subsec:data_desc}. The proportion of outliers is equal to $10\%$. For training of the One-Class SVM$+$ we use an unlabeled sample. To assess the accuracy we calculate area under the precision/recall curve using the test sample. Also we compare this accuracy with that of the original One-Class SVM. The main issue when performing comparison is how to set values for the regularization parameters and the kernel widths. Let us comment on this issue:
\begin{itemize}
\item In order to tune the regularization parameter $\nu$ and the kernel width $\sigma$ for the original One-Class SVM, we perform a grid search in order to maximize area under the precision/recall curve, estimated by the cross-validation procedure. We denote obtained ``optimal'' values by $\nu_{opt}$ and $\sigma_{opt}$ and provide the corresponding anomaly detection accuracy in figure \ref{noisy_ordinary}.   
\item For the One-Class SVM$+$ we tune only the regularization parameter $\gamma$ and the kernel width $\sigma^*$, responsible for the ``privileged'' part of optimization problem \eqref{ocsvm+}. As above, we optimize area under the precision/recall curve, estimated by the cross-validation procedure. In this case we set values of the parameters $\nu$ and $\sigma$ to $\nu_{opt}$ and $\sigma_{opt}$ correspondingly, which are optimal for the original One-Class SVM. The idea is that if privileged information does not provide any improvement over the original information, the parameter $\gamma$ can be set to some big value and  the ``privileged'' part of optimization problem \eqref{ocsvm+} will not have any influence on the overall solution. We provide typical values of the anomaly detection accuracy for the One-Class SVM+ in figure \ref{noisy}. We can see that privileged information allows obtaining significant increase of area under the precision/recall curve. 
\end{itemize}
Accuracy of anomaly detection for synthetic datasets is reported in table \ref{avprecsynth}.

\begin{table}[t!]
\centering
\begin{tabular}{|c|c|c|}
\hline
Dataset/Method & One-Class SVM\phantom{+} & One-Class SVM$+$\\
\hline
Arc & 0.25 & 0.67\\
\hline
Circles & 0.56 & 0.96\\
\hline
Mixture of Gaussians & 0.55 & 0.98\\
\hline
\end{tabular}
\caption{Synthetic datasets. Accuracy of anomaly detection}
\label{avprecsynth}
\end{table}

\subsection{Microsoft Malware Classification Challenge}
Zero-day cyber attacks such as worms and spy-ware are becoming increasingly widespread and dangerous. The existing signature-based intrusion detection mechanisms are often not sufficient in detecting these types of attacks. As a result, anomaly intrusion detection methods have been developed to cope with zero-day cyber attacks. Among the variety of common anomaly detection approaches \cite{malwa1,malwa2}, the support vector machine is known to be one of the best machine learning algorithms to classify abnormal behaviour \cite{malwa3}. In this section we demonstrate the applicability of the proposed One-Class SVM modification to cyber attacks detection using a real data from the Microsoft Malware Classification Challenge \cite{malwa}. 

In the framework of the Microsoft Malware Classification Challenge a set of known malware files, representing a mix of nine different malware families, is provided. Each malware file has an Id, a $20$ character hash value uniquely identifying the file, and a Class, an integer representing one of nine family names to which the malware may belong: Ramnit, Lollipop, Kelihos\_ver3, Vundo, Simda, Tracur, Kelihos\_ver1, Obfuscator.ACY, Gatak. For each file the raw data contains the hexadecimal representation of the file's binary content, without the PE header (to ensure sterility). A metadata manifest is also provided, which is a log containing various metadata information extracted from the binary, such as function calls, strings, etc. This was generated using the IDA disassembler tool. The task, proposed to the participants of the challenge, was to develop the best mechanism for classifying files from the test set into their respective family affiliations using binary files and assembly code.

In order to test our approach to anomaly detection with privileged information we use the same methodology of feature generation, as that initially proposed by the winning team \cite{malwa_im}. Using binary files we calculate frequencies of bytes and number of different four-grams. From the assembly code we calculate frequencies of each command, a number of calls to external dll files. Also we use transformation of the assembly code to an image, since malwares can be visualized as grayscale images from byte files or from asm files \cite{malwa_im7,malwa_im8}: each byte is from $0$ to $255$ so it can be easily translated into pixel intensity. Details of the transformation can be found in \cite{malwa_im}. Thus we use features based on image textures which are commonly used in scene category classification such as coast, mountain, forest, street,
etc. Here, instead of scene categories, we have malware families.

\begin{figure}[t!]
\begin{center}
  \includegraphics[width=.99\linewidth]{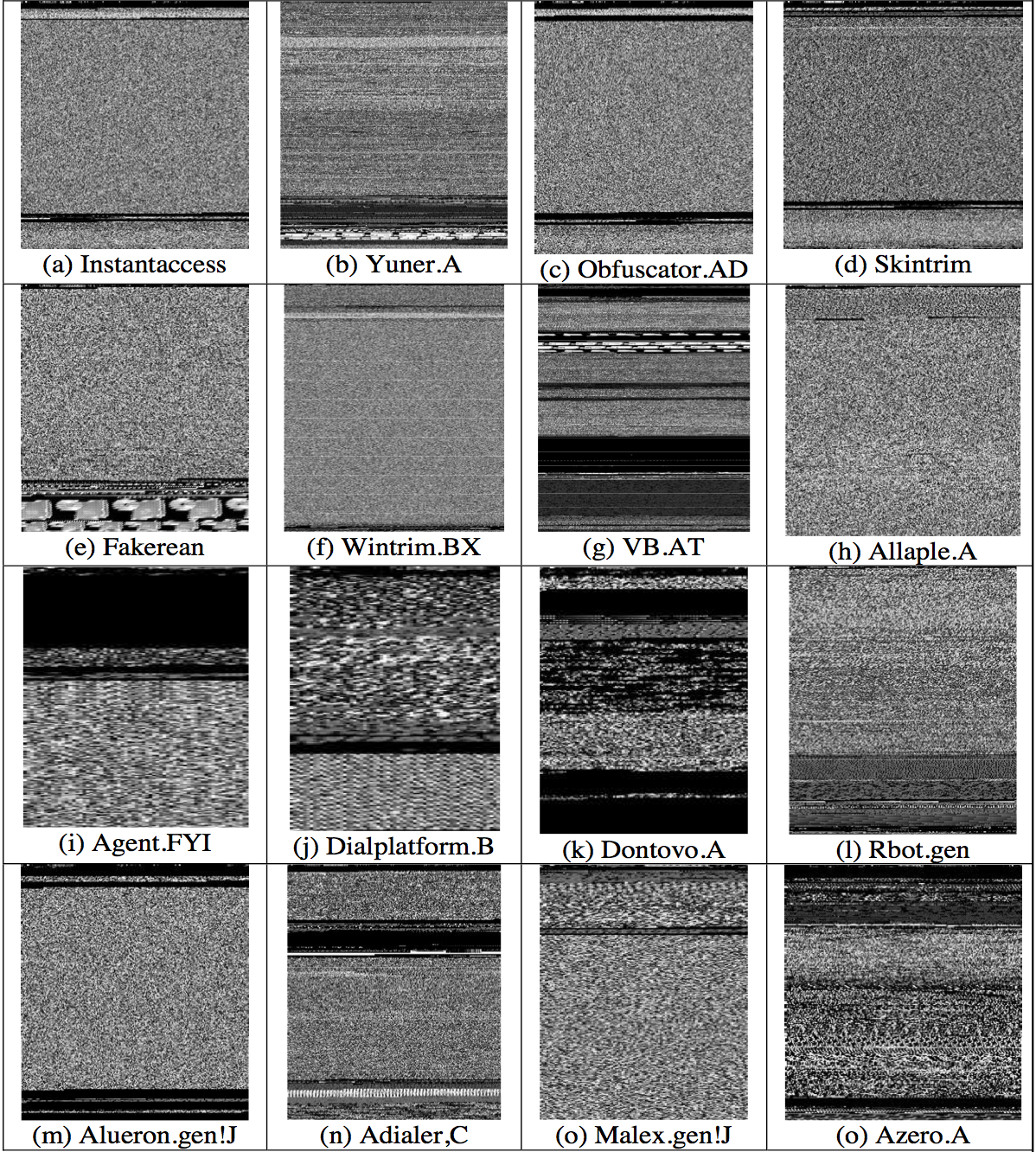}
\caption{Examples of malware assembly code transformation to images}
\label{malware_example}
\end{center}
\end{figure}

Finally, as the original features we use information, obtained from the binary files, and as privileged information we use features, obtained from the assembly code. In such a way we want to model a situation, when we have resources to perform reverse-engineering of a program in order to construct a training sample, but we can not make it during a test phase e.g. due to restrictions on computational resources.

For each of the nine malware classes we consider the following problem statement:
\begin{itemize}
\item We select one of the nine classes,
\item As a train set we use half of patterns from the selected class, 
\item As a test set we use patterns from another half of the selected class, as well as patterns from other eight classes. We consider patterns from the selected class to be ``normal'' and patterns from other classes as ``abnormal'',
\item We use  predictions on the test set to calculate area under the precision/recall curve.
\end{itemize}
We do not want to claim that such setup of experiments is indeed fully reflects a specificity of cybersecurity applications. In fact through experiments on this real malware data we would like to show that privileged information can increase model accuracy, and so our approach is useful for cyber security applications. 

As in the previous subsection, here we also perform comparison with the original One-Class SVM. In figure \ref{malware_example} we provide an example of how area under the precision/recall curve depends on the parameters. We can notice that it almost does not depend on the regularization parameter $\nu$.

Values of the parameters $\nu$ and $\sigma$, selected for the One-Class SVM by the cross-validation procedure, are also used for the One-Class SVM+. Thus, for the One-Class SVM+ we only tune values of the regularization $\gamma$ and the kernel width $\sigma^*$ in the privileged feature space. We expect that in case of a small kernel width $\sigma^*$ and a big value of $\gamma$ we obtain results similar to that of the One-Class SVM.

We provide obtained results in table \ref{malware}. For some malware classes privileged information allows getting significant increase in accuracy of anomaly detection. For other malware classes accuracies of the One-Class SVM and the One-Class SVM+ turned out to be the same thanks to the fact, that for specific values of the kernel width in the privileged feature space and specific values of the regularization parameter the decision function of the One-Class SVM+ is close to the decision function of the original One-Class SVM.

\begin{figure}[t!]
\begin{center}
\begin{subfigure}{.25\textwidth}
  \centering
  \includegraphics[width=.99\linewidth]{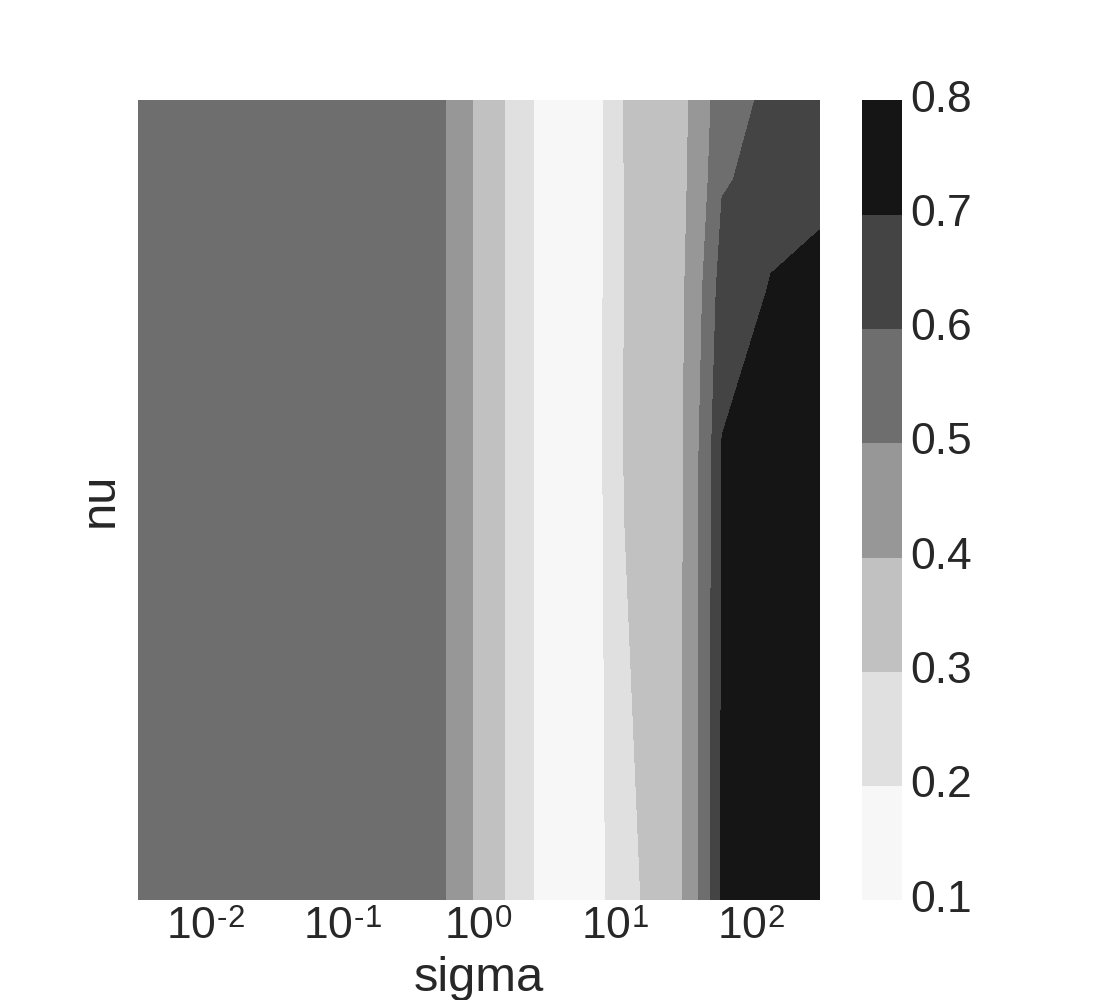}
  \caption{One-Class SVM}
\end{subfigure}%
\begin{subfigure}{.25\textwidth}
  \centering
  \includegraphics[width=.99\linewidth]{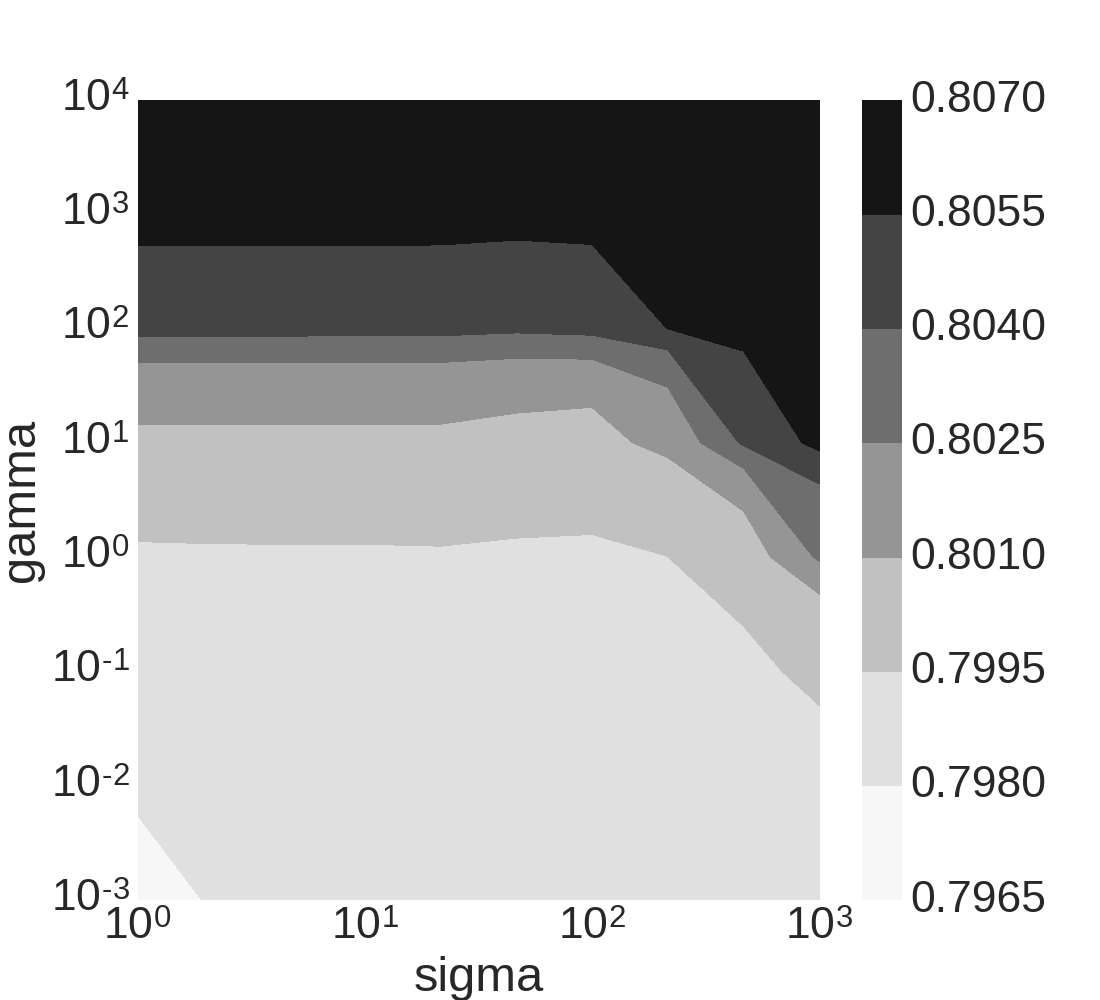}
  \caption{One-Class SVM+}
\end{subfigure} %
\caption{Malware Detection problem. Dependence of anomaly detection accuracy on the regularization parameters $\nu$ and $\gamma$, and on the Gaussian kernel widths $\sigma$ and $\sigma^*$}
\label{malware_example}
\end{center}
\end{figure}

\begin{table}[t!]
\begin{center}
\begin{tabular}{|c|c|c|c|c|c|}
\hline
Algorithm/Malware Class & 1 & 2 & 3 & 4 & 5 \\
\hline
OneClassSVM\phantom{+} & 0.67 & 0.93 & 0.97 & 0.69 & 0.57 \\
\hline
OneClassSVM+ & 0.81 & 0.95 & 0.99 & 0.72 & 0.57 \\
\hline
\end{tabular}

\begin{tabular}{c|c|c|c|c|c}
\hline
\end{tabular}

\begin{tabular}{|c|c|c|c|c|}
\hline
Algorithm/Malware Class & 6 & 7 & 8 & 9\\
\hline
OneClassSVM\phantom{+} & 0.80 & 0.82 & 0.84 & 0.60\\
\hline
OneClassSVM+ & 0.81 & 0.85 & 0.87 & 0.62\\
\hline
\end{tabular}
\caption{Malware Detection problem. Comparison of accuracies of the One-Class SVM and the One-Class SVM+}
\label{malware}
\end{center}
\end{table}

\subsection{Comparison with Related Approaches}

In \cite{ocsvmother} and \cite{svddother} (we provide the review of the related methods in section \ref{sec:otherTechniques}) authors describe results of experiments on real datasets. In particular, in  \cite{ocsvmother} results of experiments on four datasets are described, and in \cite{svddother} only two datasets are used among those, which are considered in \cite{ocsvmother}. In order to compare approaches from \cite{ocsvmother} and \cite{svddother} with methods, proposed in this paper, we use one of these two samples --- the sample Abalone. Results of experiments for other datasets are similar.

The Abalone dataset contains size and weight of molluscs, as well as their age, evaluated from a number of rings on a shell. Patterns are divided into two groups depending on the value of the parameter ``Rings''. Patterns with $\text{Rings} < 7$ are considered as a normal class, the rest patterns are considered to be anomalies.

Let us construct several blocks of privileged information.
For this we divide the dataset into two groups w.r.t. to parameters ``Length'' ($\text{Length} < 0.5$), ``Height'' ($\text{Height} < 0.15$) and	``Whole weight'' ($\text{Whole weight} < 0.8$). Each such division can be encoded by a binary vector, which we use as an additional information during the training phase. We perform three experiments. In each of the experiments we use one of these blocks of privileged information. We evaluate classification accuracy by the ten-fold cross-validation procedure.

In the previous sections we use area under the precision/recall curve in order to evaluate performance of anomaly detection algorithms. Unfortunately, in  \cite{svddother} the authors provide only accuracy values, therefore in this section for comparability we also provide only accuracy values. Results are given in table \ref{compareWithOtherImplementation}. We can see that the parametrization of the One-Class SVM+, proposed in this paper, allowed us to find more efficient solution. Also let us note that results of the One-Class SVM from \cite{ocsvmother} and of the SVDD from \cite{svddother} are comparable. 

\begin{table}
\begin{center}
\begin{tabular}{|c|c|c|c|}
\hline
Feature/Method & \begin{tabular}{@{}c@{}}One-Class \\ SVM \cite{ocsvmother}\end{tabular}   &\begin{tabular}{@{}c@{}}SVDD \\ \cite{svddother}\end{tabular} & \begin{tabular}{@{}c@{}}One-Class \\ SVM+\end{tabular} \\
\hline
Length & 0.670 & 0.673 & 0.702 \\
\hline
Height & 0.730 & 0.728 & 0.751 \\
\hline
Whole weight & 0.715 & 0.710 & 0.744 \\ 
\hline
\end{tabular}
\caption{Comparison with related approaches. Accuracy Values}
\label{compareWithOtherImplementation}
\end{center}
\end{table}

\section{Conclusions}
We provide modifications of the approaches for one-class classification problem that allows to incorporate  privileged information. We can see from the results of experiments that in some cases privileged information can significantly improve anomaly detection accuracy. In cases when privileged information is not useful for a problem at hand thanks to the structure of the corresponding optimization problem (e.g. cf. \eqref{oneclass} with \eqref{ocsvm+}) privileged information will not have a significant influence on the corresponding decision function: e.g. since for $\gamma\gg 1$ the solution of \eqref{ocsvm+} is close to the solution of \eqref{oneclass}, then the decision function of the One-Class SVM+ is close to the decision function of the original One-Class SVM.

 \section{Acknowledgements}
 \label{acknowledgement}
The research of the first author was conducted in IITP RAS and supported solely by the Russian Science Foundation grant (project 14-50-00150). The research of the second author was supported by the RFBR grants 16-01-00576 A and 16-29-09649 ofi\_m. 

\bibliographystyle{splncs}

\bibliography{references}



\end{document}